\documentclass[preprint,12pt]{elsarticle}
\usepackage[symbol]{footmisc}
\usepackage[margin=3cm]{geometry} 
\usepackage{balance,multirow,amsmath,graphicx,subcaption,hyphenat,algorithm,algpseudocode,tabularx,stfloats,booktabs}
\usepackage{verbatim} 
\newcommand{\ignore}[1]{} 

\let\today\relax
\makeatletter
\def\ps@pprintTitle{%
    \let\@oddhead\@empty
    \let\@evenhead\@empty
    \def\@oddfoot{\footnotesize\itshape
         {} \hfill\today}%
    \let\@evenfoot\@oddfoot
    }
\makeatother

\usepackage{amssymb}
\journal{N/A}

\begin{document}
\begin{frontmatter}

\title{Towards Automation of Human Stage of Decay Identification: An Artificial Intelligence Approach}

\author[inst1,inst2]{Anna-Maria Nau}
\author[inst3]{Phillip Ditto}
\author[inst3]{Dawnie Wolfe Steadman}
\author[inst1]{Audris Mockus}

\affiliation[inst1]{organization={Department of Electrical Engineering \& Computer Science, University of Tennessee},
            addressline={1520 Middle Dr}, 
            city={Knoxville},
            postcode={37996}, 
            state={TN},
            country={USA}}

\affiliation[inst2]{organization={The Bredesen Center, University of Tennessee},
            addressline={821 Volunteer Blvd.}, 
            city={Knoxville},
            postcode={37996}, 
            state={TN},
            country={USA}}

\affiliation[inst3]{organization={Department of Anthropology, University of Tennessee},
            addressline={1924 Alcoa Hwy}, 
            city={Knoxville},
            postcode={37920}, 
            state={TN},
            country={USA}}

\begin{abstract}
Determining the stage of decomposition (SOD) is crucial for estimating the postmortem interval and identifying human remains. Currently, labor-intensive manual scoring methods are used for this purpose, but they are subjective and do not scale for the emerging large-scale archival collections of human decomposition photos. This study explores the feasibility of automating two common human decomposition scoring methods proposed by Megyesi and Gelderman using artificial intelligence (AI). We evaluated two popular deep learning models, Inception V3 and Xception, by training them on a large dataset of human decomposition images to classify the SOD for different anatomical regions, including the head, torso, and limbs. Additionally, an interrater study was conducted to assess the reliability of the AI models compared to human forensic examiners for SOD identification. The Xception model achieved the best classification performance, with macro-averaged F1 scores of .878, .881, and .702 for the head, torso, and limbs when predicting Megyesi's SODs, and .872, .875, and .76 for the head, torso, and limbs when predicting Gelderman's SODs. The interrater study results supported AI's ability to determine the SOD at a reliability level comparable to a human expert. This work demonstrates the potential of AI models trained on a large dataset of human decomposition images to automate SOD identification.
\end{abstract}


\end{frontmatter}

\section{Introduction}
\label{sec:intro}
Determining the stage of decay (SOD) is an important and common task in human remains cases. Knowing the degree of decomposition is vital for estimating the postmortem interval (PMI) and identifying human remains~\cite{megyesi, gelderman, vass2011elusive, galloway1989decay}. Presently, establishing the SOD of a decedent is primarily conducted manually, via visual assessment by trained experts using non-metric scoring or staging methods, such as those proposed by Megyesi et al.~\cite{megyesi} and Gelderman et al.~\cite{gelderman}. Such non-metric methods, which rely on subjective interpretation made by humans, possess a higher susceptibility to human bias and error, consequently affecting the accuracy of downstream tasks, such as estimating the PMI~\cite{nakhaeizadeh2014cognitive,cooper2019cognitive,kukucka2017cognitive,sauerwein2018perceptions}. Furthermore, the PMI estimation formulas derived in existing studies, such as those by Megyesi et al.~\cite{megyesi} and Gelderman et al.~\cite{gelderman}, were developed using a very small number of samples. Evaluating or improving upon these formulas with a much larger sample size, such as over one million photos, would require manual SOD scoring, which is not feasible for such a large sample. Therefore, this work aims to utilize emerging artificial intelligence (AI) methods to evaluate the feasibility of automating the SOD identification task.

AI is the ability of machines to perform tasks that would typically require human intelligence and it has provided innovative approaches to assist in human decision-making~\cite{deshpande2018artificial,jarrahi2018artificial}. AI assesses information based on the entirety of acquired facts or data using advanced algorithms, thereby mitigating vulnerability to the subjectivity and biases that trouble humans and affect their decision-making abilities~\cite{piraianu2023enhancing}.  Additionally, AI algorithms can handle large volumes of data, uncovering intricate patterns that might elude human perception~\cite{deshpande2018artificial,rahmani2021artificial,korteling2021human}. This ability to process, analyze, and interpret large amounts of data quickly and precisely makes AI a valuable tool in many industries, including forensic practice and research.

In summary, the objective of this study is to evaluate the possibility of automating two established human decomposition scoring methods, namely Megyesi et al.~\cite{megyesi} and Gelderman et al.~\cite{gelderman}, using vision-based AI models, known as convolutional neural networks (CNNs). Specifically, various CNN classification models will be trained and evaluated on a large human decomposition image dataset to perform SOD prediction. In addition, an interrater test is conducted to assess and compare the reliability of the models and the human forensic examiners for SOD identification. We hypothesize that similar interrater reliability among human raters and an AI rater suggests the feasibility of using AI for SOD classification and, perhaps, other downstream tasks. The significance of such a finding lies in the potential to develop more accurate SOD and PMI estimation methods that are less effort-intensive and subjective. The primary purpose of this study is to provide a proof-of-concept for the future advancement and integration of AI-assisted analysis in forensic practice and research.

\section{Materials and methods}
\label{sec:methods}

\subsection{Stage of decay scoring methods}
\label{sec:scoring}
The two human decomposition scoring methods this study attempts to automate using AI are: (1) Megyesi et al.~\cite{megyesi} and (2) Gelderman et al.~\cite{gelderman}. To account for the differential decomposition that occurs in different body segments (e.g., limbs do not bloat or purge fluid), these two scoring methods independently assess the human body in three anatomical regions: (1) the head (including the neck), (2) the torso, and (3) the limbs (including the hands and feet). Based on the morphological features present, Megyesi et al.~\cite{megyesi} categorizes human decomposition into four high-level linear stages: fresh, early decomposition, advanced decomposition, and skeletonization. Gelderman et al.~\cite{gelderman}, building upon the work of Megyesi et al.~\cite{megyesi}, categorizes each anatomical region into six stages, with the lowest indicating no visible changes and the highest indicating complete skeletonization.

\subsection{The human decomposition dataset}
\label{sec:dataset}
The human decomposition dataset, a large-scale image collection used to train the models, includes images of decomposing corpses donated to [removed for double anonymized review]\ignore{the Forensic Anthropology Center at The University of Tennessee, Knoxville}. The center houses [removed for double anonymized review]\ignore{the Anthropology Research Facility (ARF), an outdoor decomposition laboratory}. Forensic experts from the [removed for double anonymized review]\ignore{ARF} captured these images at non-uniform intervals, with one or more days between each capture. The images, taken from various angles, depict different anatomical areas to illustrate the various stages and regions of human decomposition. The image resolutions vary from 2400×1600 pixels up to 4900×3200 pixels. The dataset covers the period from 2011 to 2023 and comprises over 1.5 million images contributed by more than 800 donors. To train different CNN classifiers on this large human decomposition image dataset to predict the SOD for various anatomical regions, the following challenges needed to be addressed:
\begin{itemize}
  \item How to best sample from the entire human decomposition dataset such that the resulting set of images reflects the dataset’s temporal characteristics. In other words, the data used to train the models should consist of images covering the entire human decay process, that is, from death to skeletonization. 
  \item Once a set of images has been sampled, the challenge is to automate the efficient extraction of specific anatomical regions (i.e., head, torso, and limbs). From a time- and cost-effective perspective, it is not feasible to manually perform this body part filtering of over one million images.
\end{itemize}
The following section (Section~\ref{sec:processing}) details how these analysis challenges were addressed during data preparation.

\subsection{Data processing and labeling}
\label{sec:processing}
\begin{figure}[h]
\centering
    \includegraphics[width=1\columnwidth]{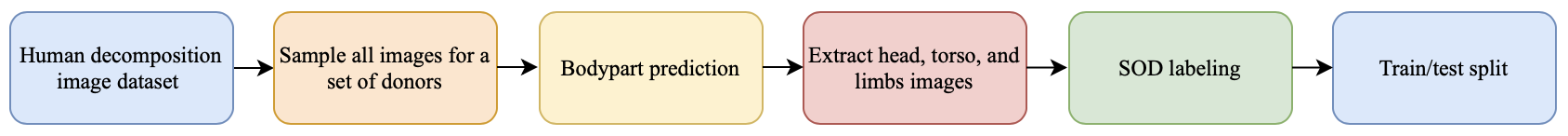}
    \caption{The data processing pipeline.}
    \label{fig:data_pipeline}
\end{figure}

The human decomposition dataset was processed according to the data pipeline shown in Figure~\ref{fig:data_pipeline}. The remainder of this section further discusses the individual steps of this pipeline.

The quality and size of the data used to train the models highly affect a model’s performance and generalizability. In other words, the more representative and diverse the training data is, the more likely it is that the model will be able to generalize. In the case of a temporal dataset, such as images documenting human decomposition, where the subjects’ appearance changes over time, it is important to sample the training data in a way that reflects the dataset’s characteristics. Therefore, images depicting all possible decomposition stages should be included in the training data. As a result, a small sample of donors, and all images of those donors, was selected over time (i.e., from when they first started to decay until fully decomposed), instead of randomly selecting images from the entire human decomposition dataset. Since this subset of images included all different kinds of anatomical areas, a previously developed body part classification model was used to automatically detect the head, torso, and limbs images to align with Megyesi et al.'s~\cite{megyesi} and Gelderman et al.'s~\cite{gelderman} scoring methods. The next step was to manually label this subset of images with the SOD labels. Specifically, the two scoring methods, described in Section~\ref{sec:scoring}, were used by a forensic expert to perform manual data labeling of the head, torso, and limbs images. Note that any body part misclassifications and/or poor quality images were either corrected or removed at this point to ensure that the final labeled datasets only included high quality images of the head, torso, and limbs. Table~\ref{table:class_labels} shows the SOD terms used in the original literature (i.e., Megyesi et al.~\cite{megyesi} and Gelderman et al.~\cite{gelderman}) and the corresponding new class labels used throughout this study. For instance, if the forensic expert, applying Megyesi et al.'s~\cite{megyesi} method, determined the SOD of a head image to be \textit{fresh}, then it was assigned the \textit{M-SOD1} class label.

\begin{table}[h!]
\centering
\begin{tabular}{ccc} 
Method & Original SOD Term & New SOD Class Label \\
\hline\hline
\\[-1em]
\multirow{4}{*}{Megyesi et al.~\cite{megyesi}} & fresh & M-SOD1 \\
    & early decomposition & M-SOD2 \\
    & advanced decomposition & M-SOD3 \\
    & skeletonization & M-SOD4 \\
\hline
\\[-1em]
\multirow{6}{*}{Gelderman et al.~\cite{gelderman}} & 1 & G-SOD1 \\
    & 2 & G-SOD2 \\
    & 3 & G-SOD3 \\
    & 4 & G-SOD4 \\
    & 5 & G-SOD5 \\
    & 6 & G-SOD6 \\
\hline\hline 
\end{tabular}
\caption{The original SOD terms mapped to the new SOD class labels used throughout this study for the different scoring methods.}
\label{table:class_labels}
\end{table}

The image labeling was conducted using an in-house developed data visualization and annotation software called [removed for double anonymized review]\ignore{ICPUTRD (Image Cloud Platform for Use in Tagging and Research on Decomposition)~\cite{nau}}. The resulting labeled datasets are shown in Table~\ref{table:datasets}, each of which will be used to train a SOD classification model. For example, \textit{M-head-data} will be used to train a model to predict the SOD (i.e., \textit{M-SOD1}, \textit{M-SOD2}, \textit{M-SOD3}, or \textit{M-SOD4}) of head images. Finally, each labeled dataset was split into a train and test set using a ratio of 80:20. The train set was used to train the model and the test set was used to evaluate the model once trained.

\begin{table}[h!]
\centering
\begin{tabular}{cccc} 
Method & Anatomical Region & Dataset & NoL \\
\hline\hline
\\[-1em]
\multirow{3}{*}{Megyesi et al.~\cite{megyesi}} & head & M-head-data & 2110 \\
    & torso & M-torso-data & 1979 \\
    & limbs & M-limbs-data & 2152 \\
\hline
\\[-1em]
\multirow{3}{*}{Gelderman et al.~\cite{gelderman}} & head & G-head-data & 2041 \\
    & torso & G-torso-data & 1982 \\
    & limbs & G-limbs-data & 2032 \\
\hline\hline 
\end{tabular}
\caption{The labeled datasets used to train the SOD classifier models. Method indicates the SOD scoring method used to label the different anatomical regions. NoL gives the number of labeled images for the particular dataset.}
\label{table:datasets}
\end{table}

\subsection{AI model development}
\label{sec:dl}
\begin{figure}[t!]
\centering
    \includegraphics[width=1\columnwidth]{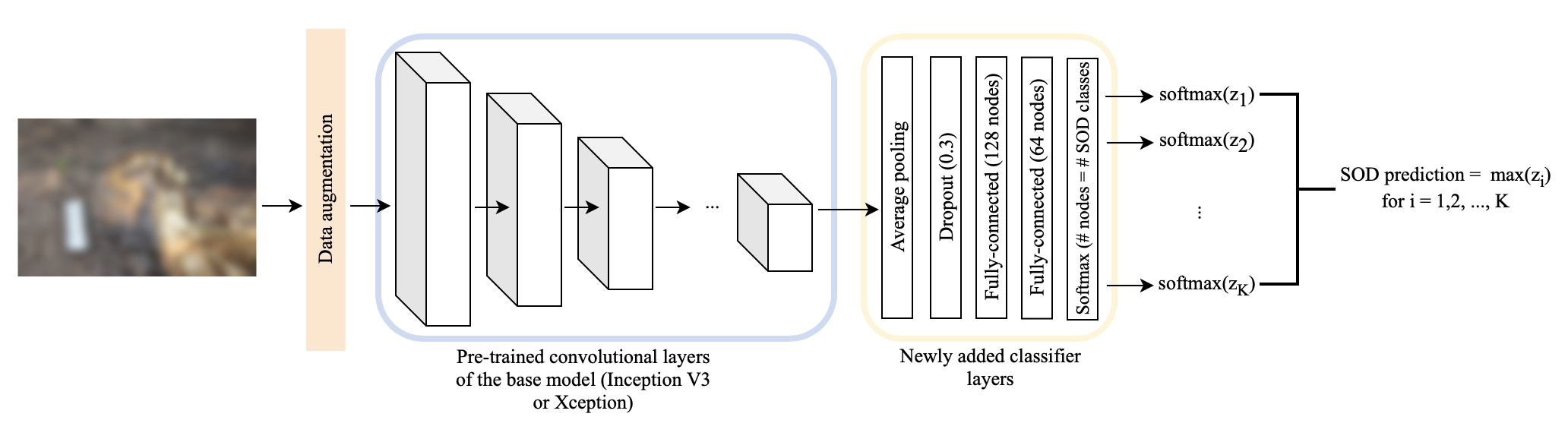}
    \caption{The SOD classification framework was trained using a two-step transfer learning approach. In step 1, the pre-trained convolutional layers of the base model (blue rectangle) were frozen, and only the newly added classifier layers (yellow rectangle) were trained. In step 2, the previously frozen layers from step 1 (blue rectangle) were unfrozen, and the entire model was trained end-to-end.}
    \label{fig:framework}
\end{figure}

To build the SOD classifiers, transfer learning was applied, which aims to produce effective models by leveraging and exploiting previously acquired knowledge~\cite{RN45}. In particular, two CNN architectures, including Inception V3~\cite{RN41} and Xception~\cite{RN40}, pre-trained on the ImageNet dataset~\cite{RN37}, were trained using the following two-step transfer learning process: (1) freeze all pre-trained convolutional layers of the base model and train newly added classifier layers and (2) unfreeze all layers and fine-tune the model end-to-end with a low learning rate. The newly added classifier layers consisted of five layers, including one global average pooling layer and one drop-out layer (rate = 0.3) to alleviate the over-fitting problem motivated by Lin et al.~\cite{lin2013network}, followed by two fully-connected layers with 128 and 64 nodes performing down-sampling, and one final softmax layer with the number of nodes equal to the number of classes for multi-class classification. More precisely, the softmax layer transforms the output of the previous layer into a probability distribution over all the classes as shown by the equation~(\ref{eqn:softmax}), with the class having the highest probability being the final SOD prediction. In equation~(\ref{eqn:softmax}), $z_i$ is the  $i^{th}$ element of the input vector to the softmax function and $K$ the total number of classes. To increase the size and diversity of the data, a data augmentation layer was added after the input layer, performing random image flipping (horizontal and vertical) and rotation during model training. Figure~\ref{fig:framework} gives an overview of the developed SOD classification framework.

\begin{equation}
\label{eqn:softmax}
    \text{softmax}(z_{i}) = \frac{e^{z_{i}}}{\sum_{j=1}^K e^{z_{j}}} \ \ \ for\ i=1,2,\dots,K
\end{equation}

The loss function used was Cross-Entropy loss, which takes the predicted probability distribution of the softmax layer and measures how well this distribution matches the true distribution. To minimize the Cross-Entropy loss function during training, the Adaptive Momentum Estimation (Adam) optimizer was employed, with a learning rate of 0.001 for the first step and 0.0001 for the second step in the two-step transfer learning process. It is worth noting that training the Inception V3 and Xception architecture from scratch (i.e., without transfer learning), transfer learning without freezing the base model (i.e., one-step transfer learning), and freezing only a certain number of base model layers instead of all followed by fine-tuning was also tested. However, the proposed two-step transfer learning process significantly improved model performances.

To evaluate the performance of the trained SOD classifier models, the confusion matrix on the test data was calculated per class, which summarizes a model's performance by comparing its predicted labels to its true labels. Specifically, the confusion matrix shows the number of correct predictions, such as the true positives (TP) and the true negatives (TN), as well as the number of incorrect predictions, such as the false positives (FP) and the false negatives (FN). The following two performance metrics were then calculated from the per-class confusion matrix: precision~(\ref{eqn:precision}) and recall~(\ref{eqn:recall}). Precision measures the accuracy of the positive predictions, while recall measures the completeness of the positive predictions.

\begin{equation}
\label{eqn:precision}
    \text{precision} = \frac{\text{TP}}{\text{TP+FP}}
\end{equation}

\begin{equation}
\label{eqn:recall}
    \text{recall} = \frac{\text{TP}}{\text{TP+FN}}
\end{equation}

To combine the per-class precision and recall metrics into a single model evaluation metric, the macro-averaged F1 score (mF1)~(\ref{eqn:mf1}) was calculated and reported, which is the unweighted mean of the per-class F1 scores~(\ref{eqn:f1}).

\begin{equation}
\label{eqn:mf1}
        \text{mF1} = \frac{\sum_{i=1}^n \text{F1}_i}{n} \ \ \ for\ i=1,2,\dots,n, \text{where $n$ is the number of classes}
\end{equation}

\begin{equation}
\label{eqn:f1}
    \text{F1} = \frac{2*(\text{Precision}*\text{Recall})}{\text{Precision}+\text{Recall}}
\end{equation}

\subsection{Interrater test} 
\label{sec:reliability_study}
To ensure that manual rating techniques are reliable, interrater reliability tests are often used to assess how similar the ratings are between two or more raters on the same set of data samples. In this study, it is also used to assess the reliability of the developed models. The interrater test involved multiple raters (both human and AI) labeling a set of images depicting the same anatomical region using Megyesi et al.'s~\cite{megyesi} and Gelderman et al.'s~\cite{gelderman} scoring methods. Specifically, 300 torso images that were not used during model development were selected. Due to limited resources, the interrater test focused on the torso only, which provides a good amount of variability. The raters included the developed torso SOD model or \textit{Model} and three forensic experts well-versed in the considered scoring methods, including \textit{Human 1} (the same human who labeled the data the models were trained with), \textit{Human 2}, and \textit{Human 3}. The task for each rater was to label the 300 images using once Megyesi et al.'s~\cite{megyesi} method and once Gelderman et al.'s~\cite{gelderman} method. The \textit{Model} rater performed labeling by predicting the SOD of the 300 images, while the human raters were instructed to independently label the 300 images on [removed for double anonymized review]\ignore{ICPUTRD~\cite{nau}}, following a similar set-up used for data labeling in Section~\ref{sec:processing}. To ensure randomization across methods, human labeling was conducted in batches of images instead of labeling all 300 images with one method and then the other. Specifically, the human raters were presented with batches of 50 images at a time, which they were asked to label using one scoring method (i.e., Megyesi et al.~\cite{megyesi} or Gelderman et al.~\cite{gelderman}). This process was repeated until all 300 images were labeled with both scoring methods, resulting in 12 iterations for each human rater ((300 images / 50 images) × 2 methods). Note, the method used to label a batch of images was alternated between the two scoring methods. 

After each rater completed labeling, two types of agreements were assessed: (1) human-human agreement (i.e., the agreement among all three human raters: \textit{Human 1}, \textit{Human 2}, and \textit{Human 3}) and (2) AI-human agreement (i.e., the agreement among the \textit{Model} replacing \textit{Human 1}, and the other two human raters, \textit{Human 2}, and \textit{Human 3}). The reason for the \textit{Model} replacing \textit{Human 1} was to see how the agreement changes when the human is replaced by the model trained on the data they labeled. To measure the different agreements, the Fleiss’ Kappa statistic was used, which measures the degree of agreement among raters over what would be expected by random chance, with values ranging from -1 (no agreement) to 1 (perfect agreement). The Fleiss’ Kappa values will be interpreted based on Landis and Kock’s~\cite{landis1977measurement} interpretation criteria shown in Table~\ref{table:kappa_evaluation}.

\begin{table}[h!]
\centering
\begin{tabular}{cc} 
Fleiss' Kappa & Level of Agreement \\
\hline\hline
    $\geq$ 0.8 & almost perfect \\
    $\geq$ 0.6 & substantial \\
    $\geq$ 0.4 & moderate \\
    $\geq$ 0.2 & fair \\
    $\geq$ 0 & slight \\
    $<$ 0 & no agreement \\
\hline\hline
\end{tabular}
\caption{Landis and Kock’s~\cite{landis1977measurement} Fleiss' Kappa interpretation criteria.}
\label{table:kappa_evaluation}
\end{table}

\section{Results}
\label{sec:results}
\subsection{SOD classification}
\label{sec:classification_results}
The SOD classifier models were implemented using Keras and TensorFlow, two open-source machine learning modules written in Python. In particular, the models were built and evaluated on the six manually labeled datasets shown in Table~\ref{table:datasets}. All images were resized to 299×299 pixels, as that is the required input image size for both the Inception V3 and Xception architectures. All models were trained on a single Tesla V100\hyp{}SXM2 GPU with 32GB of memory. The batch size was set to 32, and the number of epochs was set to 200, with early stopping set to 20 epochs to avoid over-fitting on the training set.

Table~\ref{table:megyesi_results} presents the Megyesi et al.~\cite{megyesi} SOD classification results on the test data. For each anatomical region, the best SOD classification performance, as indicated by the mF1 scores, was achieved with the Xception architecture. The head SOD classifier achieved an mF1 score of .878, and the torso SOD classifier achieved an mF1 score of .881, performing comparably. However, the limbs SOD classifier performed notably less well, with an mF1 score of .702.

\setlength{\tabcolsep}{7pt} 
\begin{table}[h!]
\centering
    \begin{tabular}{cc cccccccc c} 
    \multirow{3}{1.9cm}{Archt.} & \multirow{3}{*}{Region} & \multicolumn{8}{c}{Per-Class Precision (P) \& Recall (R)} & \\ 
    \cline{3-10}
    && \multicolumn{2}{c}{M-SOD1} & \multicolumn{2}{c}{M-SOD2} & \multicolumn{2}{c}{M-SOD3} & \multicolumn{2}{c}{M-SOD4} & \multirow{3}{*}{mF1} \\  
    \cline{3-10}
    && P & R & P & R & P & R & P & R & \\
    \hline\hline
    \multirow{3}{1.9cm}{InceptionV3} 
                                & head & .846 & .805 & .835 & .91 & .698 & .677 & .882 & .778 & .806 \\
                                & torso & 1.0 & 1.0 & .936 & .83 & .683 & .707 & .733 & .892 & .845 \\
                                & limbs & .667 & .333 & .94 & .94 & .467 & .5 & .897 & .929 & .695 \\  
    \hline
    \multirow{3}{1.9cm}{Xception} 
                                & head & .932 & .842 & .878 & .966 & .887 & .723 & .893 & .926 & .878 \\
                                & torso & 1.0 & 1.0 & .926 & .898 & .754 & .767 & .829 & .872 & .881 \\
                                & limbs & .75 & .5 & .924 & .978 & .6 & .214 & .92 & .967 & .702 \\
    \hline\hline
    \end{tabular}
\caption{Megyesi et al.'s~\cite{megyesi} SOD classification results.}
\label{table:megyesi_results}
\end{table}

Table~\ref{table:gelder_results} presents the Gelderman et al.~\cite{gelderman} SOD classification results on the test data.  Similar to the Megyesi et al.~\cite{megyesi} results, the best classification performance, as indicated by the mF1 scores, was achieved by the Xception architecture across all anatomical regions. The head SOD classifier achieved an mF1 score of .872, and the torso SOD classifier achieved an mF1 score of .875, performing comparably. However, the limbs SOD classifier performed less well, with an mF1 score of .76. 

\setlength{\tabcolsep}{2.4pt} 
\begin{table}[h!]
\centering
    \begin{tabular}{cc cccccccccccc c} 
    \multirow{3}{2cm}{Archt.} & \multirow{3}{*}{Region} & \multicolumn{12}{c}{Per-Class Precision (P) \& Recall (R)} & \\ 
    \cline{3-14}
    && \multicolumn{2}{c}{G-SOD1} & \multicolumn{2}{c}{G-SOD2} & \multicolumn{2}{c}{G-SOD3} & \multicolumn{2}{c}{G-SOD4} & \multicolumn{2}{c}{G-SOD5} & \multicolumn{2}{c}{G-SOD6} & \multirow{3}{*}{mF1} \\  
    \cline{3-14}
    && P & R & P & R & P & R & P & R & P & R & P & R &\\
    \hline\hline
    \multirow{3}{2cm}{InceptionV3} 
                                & head & 1.0 & .8 & .903 & .933 & .946 & .907 & .8 & .8 & .696 & .8 & .895 & .944 & .866 \\
                                & torso & .2 & .5 & .862 & .862 & .954 & .954 & .897 & .833 & .7 & .824 & .833 & .714 & .749 \\
                                & limbs & .75 & .75 & .704 & 1.0 & .907 & .739 & .696 & .765 & .512 & 1.0 & 1.0 & 0.059 & .651 \\   
    \hline
    \multirow{3}{2cm}{Xception} 
                                & head & 1.0 & .6 & .882 & 1.0 & .947 & .918 & .806 & .829 & .818 & .9 & 1.0 & .889 & .872 \\
                                & torso & .667 & 1.0 & .964 & .931 & .974 & .862 & .792 & 1.0 & .789 & .81 & .944 & .81 & .875 \\
                                & limbs & 1.0 & 1.0 & .875 & .947 & .909 & .87 & .75 & .824 & .826 & .905 & 1.0 & 0.071 & .76 \\
    \hline\hline
    \end{tabular}
\caption{Gelderman et al.'s~\cite{gelderman} SOD classification results.}
\label{table:gelder_results}
\end{table}

\subsection{Interrater test}
\label{sec:Interrater_results}
The interrater test results are shown in Table~\ref{table:interrater_results}. Specifically, a Fleiss' Kappa analysis using SPSS Statistics was conducted for both the human-human agreement and AI-human agreement across both scoring methods. The reported agreement levels were determined using Landis and Kock’s~\cite{landis1977measurement} interpretation criteria, as shown in Table~\ref{table:kappa_evaluation}. According to the Megyesi et al.~\cite{megyesi} results, the Fleiss' Kappa coefficient of the human-human agreement was .67 with a p-value $<$ .001 and a 95\% confidence interval (CI) of .628 to .713, indicating \textit{substantial} agreement. Similarly, the Fleiss' Kappa coefficient of the AI-human agreement was .637 with a p-value $<$ .001 and a 95\% CI of .594 to .68, suggesting \textit{substantial} agreement. Additionally, according to the Gelderman et al.~\cite{gelderman} results, the Fleiss' Kappa coefficient of the human-human agreement was .593 with a p-value $<$ .001 and a 95\% CI of .558 to .628, indicating \textit{moderate} agreement. Similarly, the Fleiss' Kappa coefficient of the AI-human agreement was .558 with a p-value $<$ .001 and a 95\% CI of .524 to .592, suggesting \textit{moderate} agreement. In all cases, the Fleiss' Kappa coefficient was statistically significant (i.e., the p-value $<$ 0.05). 

\setlength{\tabcolsep}{4pt} 
\begin{table}[h!]
\centering
\begin{tabular}{ccccccc} 
\multirow{2}{*}{Method} & \multirow{2}{*}{Agreement} & \multirow{2}{*}{Fleiss' Kappa} & \multirow{2}{*}{P-value} & \multicolumn{2}{c}{95\% CI} & Agreement \\ 
\cline{5-6}
&&&& UB & LB & Level\\
\hline\hline
\multirow{2}{*}{Megyesi et al.~\cite{megyesi}} & human-human & .67 & $<$.001 & .628 & .713 &  substantial \\
                        & AI-human &  .637 & $<$.001 & .594 & .68 & substantial \\
\hline
\\[-1em]
\multirow{2}{*}{Gelderman et al.~\cite{gelderman}} & human-human & .593 & $<$.001 & .558 & .628 & moderate \\
                            & AI-human &  .558 & $<$.001 & .524 & .592 & moderate \\
                                                    
\hline\hline 
\end{tabular}
\caption{The interrater study results. Reported are the Fleiss' kappa value, the p-value, the upper bound (UB) and lower bound (LB) of the 95\% confidence interval (CI), and the agreement level for both the human-human and AI-human agreement across both scoring methods. The agreement levels are determined using Landis and Kock’s~\cite{landis1977measurement} interpretation criteria, as shown in Table~\ref{table:kappa_evaluation}.}
\label{table:interrater_results}
\end{table}

\subsection{Discussion}
\label{sec:discussion}
Overall, the SOD classification results are promising. The Xception architecture performed the best across both scoring methods. The head and torso SOD models performed comparably well; however, the limbs SOD models' performances were not as strong. Further analysis of the limb data revealed that some images included hands and/or feet covered by a net (to prevent them from being scattered/disarticulated by animal scavengers), which could confuse and distract the model, leading to incorrect predictions. Additionally, the examination of the limbs data indicated that some images included other parts of the body, specifically the torso. Since the torso decays differently than the limbs, this inclusion could again confuse the model and, consequently, affect its predictions. Future work will focus on addressing these data quality challenges to ensure the development of a more reliable limbs dataset and hence improved prediction performance. Another important finding to mention is that although the Gelderman et al.~\cite{gelderman} datasets contained more SOD classes (six classes) than the Megyesi et al.~\cite{megyesi} datasets (four classes), the classification performances were comparable. This indicates that these AI models are able to learn a higher number of decay stages without decreasing prediction performance.

While the SOD prediction performances were promising overall, there is room for improvement, as indicated by the per-class precision and recall values. The sizes of the labeled datasets used in this study are considered rather small for training deep learning architectures, such as Inception V3 and Xception. Additionally, a deeper analysis of the labeled datasets indicated class imbalance (i.e., a disproportionate number of instances of one class compared to another). Having a larger, more diverse, and evenly balanced dataset will make the models more robust and improve their generalization capabilities, which refers to how well a model can react to new and unseen data. However, creating more labeled data where domain expertise is required may be limited by both resource and time constraints. A recent study [removed for double anonymized review]\ignore{by Nau et al.~\cite{nau2023stage}} addressed such challenges by developing a domain-aware label propagation algorithm that leverages different image attributes to automatically perform data labeling, thereby reducing manual labeling efforts and costs. Future work will explore integrating such label propagation methods to obtain larger and more diverse datasets, aiming to create more robust and accurate SOD classification models.

In the interrater test, both the human-human and AI-human agreements showed \textit{substantial} agreement when applying the Megyesi et al.~\cite{megyesi} scoring method. Using the Gelderman et al.~\cite{gelderman} scoring method, \textit{moderate} agreement was observed for both the human-human and AI-human agreements. Across both scoring methods, the level of agreement for both the human-human and AI-human agreements was the same. This means that when the human rater was replaced with the AI model rater, the level of reliability stayed the same, supporting AI's ability to perform SOD identification with a reliability level comparable to that of an experienced human forensic examiner. Notably, the lower agreement level for the Gelderman et al.~\cite{gelderman} method could be attributed to (1) its novelty in the field and/or (2) its complexity, being more complicated to apply, as it involves six decay stages compared to the four stages in the Megyesi et al.~\cite{megyesi} method.

While the results demonstrate that human decomposition scoring methods have the potential to be automated using AI techniques, there are some important limitations of this work. For one, the data labeling was conducted by a single forensic expert. This approach may introduce labeling bias, which can lead to inherently biased training datasets. Models trained on such datasets can inherit these biases, resulting in biased models~\cite{jiang2020identifying}. Therefore, future work will focus on creating a so-called ``gold standard" dataset—a labeled dataset meticulously crafted and evaluated by multiple forensic experts. Such a dataset would be accepted as the most accurate and reliable of its kind. This step will ensure that the models are trained with accurate and unbiased data, which is vital for developing high-quality models. 

An additional limitation is that this study is environmental- and climate-specific. The images used to train the models were all taken of donors decaying outdoors in an open-wooded area with the ground consisting of soil, gravel, and dead/decaying plant matter (e.g., rotting wood and shedding leaves). Additionally, the climate of this area is humid subtropical, characterized by high summer and moderate winter temperatures. Therefore, the models may not perform as well on images taken in different climate conditions or environments, necessitating additional training or re-training with images specific to those conditions.

\section{Conclusion}
\label{sec:Conclusion}
This study explored the possibility of automating two common human decomposition scoring methods, namely Megyesi et al.~\cite{megyesi} and Gelderman et al.~\cite{gelderman}. Specifically, different CNN models, including Inception V3 and Xception, were trained on a large human decomposition image dataset to classify the SOD for different anatomical regions. Across both scoring methods, the Xception model achieved the highest classification results, performing comparably well for the head and torso, and slightly lower for the limbs. The interrater reliability study results provided support for AI's ability to automate the SOD identification task at a reliability level comparable to a human expert. Overall, the study results are promising and provide a proof-of-concept for automating human decomposition scoring methods using AI.

\bibliographystyle{elsarticle-num} 
\bibliography{references}

\begin{thebibliography}{10}
\expandafter\ifx\csname url\endcsname\relax
  \def\url#1{\texttt{#1}}\fi
\expandafter\ifx\csname urlprefix\endcsname\relax\def\urlprefix{URL }\fi
\expandafter\ifx\csname href\endcsname\relax
  \def\href#1#2{#2} \def\path#1{#1}\fi

\bibitem{megyesi}
M.~S. Megyesi, S.~P. Nawrocki, N.~H. Haskell, Using accumulated degree-days to estimate the postmortem interval from decomposed human remains, Journal of Forensic Science 50~(3) (2005) 618--626.
\newblock \href {https://doi.org/10.1.1/jpb001} {\path{doi:10.1.1/jpb001}}.

\bibitem{gelderman}
H.~Gelderman, L.~Boer, T.~Naujocks, A.~IJzermans, W.~Duijst, The development of a post-mortem interval estimation for human remains found on land in the netherlands, International journal of legal medicine 132~(3) (2018) 863--873.
\newblock \href {https://doi.org/10.1007/s00414-017-1700-9} {\path{doi:10.1007/s00414-017-1700-9}}.

\bibitem{vass2011elusive}
A.~A. Vass, The elusive universal post-mortem interval formula, Forensic science international 204~(1-3) (2011) 34--40.
\newblock \href {https://doi.org/10.1016/j.forsciint.2010.04.052} {\path{doi:10.1016/j.forsciint.2010.04.052}}.

\bibitem{galloway1989decay}
A.~Galloway, W.~H. Birkby, A.~M. Jones, T.~E. Henry, B.~O. Parks, Decay rates of human remains in an arid environment, Journal of forensic sciences 34~(3) (1989) 607--616.
\newblock \href {https://doi.org/10.1520/JFS12680J} {\path{doi:10.1520/JFS12680J}}.

\bibitem{nakhaeizadeh2014cognitive}
S.~Nakhaeizadeh, I.~E. Dror, R.~M. Morgan, Cognitive bias in forensic anthropology: visual assessment of skeletal remains is susceptible to confirmation bias, Science \& Justice 54~(3) (2014) 208--214.
\newblock \href {https://doi.org/10.1016/j.scijus.2013.11.003} {\path{doi:10.1016/j.scijus.2013.11.003}}.

\bibitem{cooper2019cognitive}
G.~S. Cooper, V.~Meterko, Cognitive bias research in forensic science: a systematic review, Forensic science international 297 (2019) 35--46.
\newblock \href {https://doi.org/10.1016/j.forsciint.2019.01.016} {\path{doi:10.1016/j.forsciint.2019.01.016}}.

\bibitem{kukucka2017cognitive}
J.~Kukucka, S.~M. Kassin, P.~A. Zapf, I.~E. Dror, Cognitive bias and blindness: a global survey of forensic science examiners., Journal of applied research in memory and cognition 6~(4) (2017) 452--459.
\newblock \href {https://doi.org/10.1016/j.jarmac.2017.09.001} {\path{doi:10.1016/j.jarmac.2017.09.001}}.

\bibitem{sauerwein2018perceptions}
K.~A. Sauerwein, Perceptions and cognitive bias in decomposition scoring methods in forensic anthropology, Ph.D. thesis, University of Tennessee (2018).

\bibitem{deshpande2018artificial}
A.~Deshpande, M.~Kumar, Artificial intelligence for big data: complete guide to automating big data solutions using artificial intelligence techniques, Packt Publishing Ltd, 2018.

\bibitem{jarrahi2018artificial}
M.~H. Jarrahi, Artificial intelligence and the future of work: human-ai symbiosis in organizational decision making, Business horizons 61~(4) (2018) 577--586.
\newblock \href {https://doi.org/10.1016/j.bushor.2018.03.007} {\path{doi:10.1016/j.bushor.2018.03.007}}.

\bibitem{piraianu2023enhancing}
A.-I. Piraianu, A.~Fulga, C.~L. Musat, O.-R. Ciobotaru, D.~G. Poalelungi, E.~Stamate, O.~Ciobotaru, I.~Fulga, Enhancing the evidence with algorithms: how artificial intelligence is transforming forensic medicine, Diagnostics 13~(18) (2023) 2992--3003.
\newblock \href {https://doi.org/10.3390/diagnostics13182992} {\path{doi:10.3390/diagnostics13182992}}.

\bibitem{rahmani2021artificial}
A.~M. Rahmani, E.~Azhir, S.~Ali, M.~Mohammadi, O.~H. Ahmed, M.~Y. Ghafour, S.~H. Ahmed, M.~Hosseinzadeh, Artificial intelligence approaches and mechanisms for big data analytics: a systematic study, PeerJ computer science 7 (2021) 1--28.
\newblock \href {https://doi.org/10.7717/peerj-cs.488} {\path{doi:10.7717/peerj-cs.488}}.

\bibitem{korteling2021human}
J.~H. Korteling, G.~C. van~de Boer-Visschedijk, R.~A. Blankendaal, R.~C. Boonekamp, A.~R. Eikelboom, Human-versus artificial intelligence, Frontiers in artificial intelligence 4 (2021) 1--13.
\newblock \href {https://doi.org/10.3389/frai.2021.622364} {\path{doi:10.3389/frai.2021.622364}}.

\bibitem{RN45}
S.~J. Pan, Q.~Yang, A survey on transfer learning, IEEE Transactions on knowledge and data engineering 22~(10) (2010) 1345--1359.
\newblock \href {https://doi.org/10.1109/TKDE.2009.191} {\path{doi:10.1109/TKDE.2009.191}}.

\bibitem{RN41}
C.~Szegedy, V.~Vanhoucke, S.~Ioffe, J.~Shlens, Z.~Wojna, Rethinking the inception architecture for computer vision, in: IEEE conference on computer vision and pattern recognition, 2016, pp. 2818--2826.
\newblock \href {https://doi.org/10.1109/CVPR.2016.308} {\path{doi:10.1109/CVPR.2016.308}}.

\bibitem{RN40}
F.~Chollet, Xception: deep learning with depthwise separable convolutions, in: IEEE conference on computer vision and pattern recognition, 2017, pp. 1251--1258.
\newblock \href {https://doi.org/10.48550/arXiv.1610.02357} {\path{doi:10.48550/arXiv.1610.02357}}.

\bibitem{RN37}
J.~Deng, W.~Dong, R.~Socher, L.-J. Li, K.~Li, L.~Fei-Fei, Imagenet: a large-scale hierarchical image database, in: IEEE conference on computer vision and pattern recognition, 2009, pp. 248--255.
\newblock \href {https://doi.org/10.1109/CVPR.2009.5206848} {\path{doi:10.1109/CVPR.2009.5206848}}.

\bibitem{lin2013network}
M.~Lin, Q.~Chen, S.~Yan, Network in network, arXiv preprint arXiv:1312.4400 (2013).
\newblock \href {https://doi.org/10.48550/arXiv.1312.4400} {\path{doi:10.48550/arXiv.1312.4400}}.

\bibitem{landis1977measurement}
J.~R. Landis, G.~G. Koch, The measurement of observer agreement for categorical data, Biometrics (1977) 159--174\href {https://doi.org/10.2307/2529310} {\path{doi:10.2307/2529310}}.

\bibitem{jiang2020identifying}
H.~Jiang, O.~Nachum, Identifying and correcting label bias in machine learning, in: International conference on artificial intelligence and statistics, PMLR, 2020, pp. 702--712.
\newblock \href {https://doi.org/10.48550/arXiv.1901.04966} {\path{doi:10.48550/arXiv.1901.04966}}.

\end{thebibliography}

\end{document}